\documentclass[
twocolumn,
]{ceurart}

\usepackage{listings}
\usepackage{todonotes}
\usepackage{float}
\begin{document}

\copyrightyear{2023}
\copyrightclause{Copyright for this paper by its authors.
  Use permitted under Creative Commons License Attribution 4.0
  International (CC BY 4.0).}

\conference{RecSys in HR'23: The 3rd Workshop on Recommender Systems for Human Resources, in conjunction with the 17th ACM Conference on Recommender Systems, September 18--22, 2023, Singapore, Singapore.}

\title{Career Path Recommendations for Long-term Income Maximization: A Reinforcement Learning Approach}


\author[1]{Spyros Avlonitis}[email=spyrosavl@gmail.com]
\fnmark[1]
\address[1]{KLM Royal Dutch Airlines, Amstelveen, The Netherlands}
\address[2]{Meta, Amsterdam, The Netherlands}
\address[3]{University of Amsterdam, Amsterdam, The Netherlands}
\address[4]{Discovery Lab, Elsevier, Amsterdam, The Netherlands}
\address[5]{Randstad, Diemen, The Netherlands}
\author[2]{Dor Lavi}[email=dor.la.vie@gmail.com]
\author[3,4]{Masoud Mansoury}[email=m.mansoury@uva.nl]
\author[5]{David Graus}[email=david.graus@randstadgroep.nl]

\fntext[1]{Work done while on internship at Randstad Groep Nederland.}

\begin{abstract}
This study explores the potential of reinforcement learning algorithms to enhance career planning processes. Leveraging data from Randstad The Netherlands, the study simulates the Dutch job market and develops strategies to optimize employees' long-term income. 
By formulating career planning as a Markov Decision Process (MDP) and utilizing machine learning algorithms such as Sarsa, Q-Learning, and A2C, we learn optimal policies that recommend career paths with high-income occupations and industries. The results demonstrate significant improvements in employees' income trajectories, with RL models, particularly Q-Learning and Sarsa, achieving an average increase of 5\% compared to observed career paths. 
The study acknowledges limitations, including narrow job filtering, simplifications in the environment formulation, and assumptions regarding employment continuity and zero application costs. 
Future research can explore additional objectives beyond income optimization and address these limitations to further enhance career planning processes.
\end{abstract}


\maketitle

\section{Introduction}
The importance of career planning in shaping an individual's professional journey cannot be overstated. It involves strategic decision-making related to one's career goals, which may be as diverse as the individuals themselves. However, despite varied ambitions, a proactive approach towards career planning universally benefits all, allowing individuals to align their career trajectory with their objectives, such as maximizing lifetime income. Recognizing that reality often presents multifaceted goals and constraints, this paper aims to simplify the career planning process using the power of artificial intelligence.

The efficacy of career planning significantly depends on the insight one has into potential career paths and their expected rewards. This study leverages AI to provide such insights to employees. Collaborating with Randstad, a global leader in the HR services industry, this paper harnesses a vast array of data encompassing anonymized employee profiles, job applications, and salary information. Using machine learning, we simulate the Dutch job market and employ reinforcement learning to strategize for maximizing employees' long-term income. Although income is not the sole objective for everyone, this research assumes it as the sole optimizing objective for simplicity. However, the proposed framework is flexible and can accommodate other objectives, such as job satisfaction or a mix of objectives, given the availability of relevant data.

The primary goal is to design a system that uses an employee's work experience as input to recommend a career path, a series of occupations and industries, that on average, delivers the highest income over ten years. A ten-year timescale is selected under the assumption that job market dynamics remain unpredictable beyond this period, making further recommendations potentially unreliable. It is important to note that the suggested career paths should be practical, implying a high likelihood of hiring should employees opt to pursue them.

\section{Background}

In this section, we briefly describe general reinforcement learning architecture and review the literature on career path recommendations.

\subsection{Reinforcement Learning}
Reinforcement Learning (RL), as characterized by \citet{Sutton2018ReinforcementIntroduction}, is a decision-making paradigm adept at handling tasks with potential delayed outcomes, such as career planning. Unlike other learning strategies, RL operates on trial-and-error, aiming to optimize a specific metric without any direct instruction. It involves an agent navigating an environment to maximize a cumulative reward over time. The RL system incorporates six primary elements: the \textbf{Agent} that interacts with the environment based on its policy, the \textbf{Environment} providing feedback, the \textbf{State} and \textbf{Action} representing the environment, and the choices available, the \textbf{Reward} as a numerical feedback, and the \textbf{Policy} directing the agent's actions.

\subsubsection{Markov Decision Processes}
The application of RL to career planning necessitates formulating the problem as a Markov Decision Process (MDP), as suggested by Puterman \cite{puterman1990markov}. This enables us to leverage established RL research and precise theoretical results. MDPs formalize sequential decision-making where actions influence not only immediate rewards but also future states, and by extension, future rewards. The inherent Markov property in an MDP posits that the transition probabilities to a new state depend solely on the current state and action.
    
\begin{figure}
    \centering
    \includegraphics[width=\columnwidth]{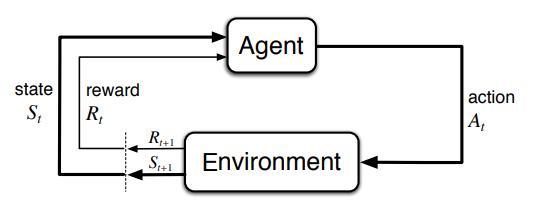}
    \caption{The agent-environment interaction in a Markov decision process.}
    \label{fig:rl_mdps}
\end{figure}

\subsection{Recommender Systems in Human Resources}

Historically, research into workforce mobility and career development has utilized traditional data sources such as surveys and censuses, as noted by \citet{Topel1992JobMen} and \citet{Long2006LabourHistory}. 
However, the rise of Online Professional Networks (OPN) has allowed for the employment of data-driven machine learning methods. The focus has increasingly shifted towards modelling career paths to predict mobility and aid in career development. This has proved valuable for both employers and employees, facilitating strategic decision-making in hiring and career progression.
    
Several studies have taken varied approaches to this issue. For instance, \citet{Paparrizos2011MachineRecommendation} employed a naive Bayes model to predict job transitions, while \citet{Wang2013IsSwitch} used a proportional hazards model to estimate when employees might decide to change jobs. Further, \citet{Liu2016FortunePath} explored career path prediction using social network data, while \citet{Li2017NEMO:Embedding} introduced the NEMO model for predicting future company and job titles using Long Short-Term Memory (LSTM) networks.
    
The advent of more complex models has also been witnessed. \citet{Meng2019APrediction} used a hierarchical neural network with an embedded attention mechanism, and \citet{Xu2019DynamicModeling} performed a talent flow analysis for predicting the increments in a dynamic job transition network. Other models, like the one proposed by \citet{Liu2020TowardsPrediction}, utilized logistic regression to predict career choices, while \citet{Al-Dossari2020} proposed a recommendation system for IT graduates based on skill similarity.
    
A separate line of research rejects the notion that frequently observed paths are necessarily the most beneficial. 
\citet{Lou2010APlanning} recommended the shortest career path using a Markov Chain model, whereas \citet{Oentaryo2018JobComposerLearning} focused on achieving the best payoff trade-off in career path planning. 
\citet{Shahbazi2019OptimizationOpportunities} optimized towards the career development of employees rather than productivity. Other approaches have included the use of skill graphs for transition pathway recommendations as demonstrated by \citet{Gugnani2019GeneratingRecommendation} and \citet{Dawson2021Skill-drivenPathways}, and the use of reinforcement learning for dynamic career path recommendations as presented by \citet{Kokkodis2021Demand-awareApproach}. 
Most recently, \citet{Guo2022IntelligentLearning} proposed a reinforcement learning variant for optimizing career paths.

The research presented in this paper is similar to previous work such as that of \citet{Oentaryo2018JobComposerLearning}, \citet{Kokkodis2021Demand-awareApproach}, and \citet{Guo2022IntelligentLearning}. 
Unlike \citet{Kokkodis2021Demand-awareApproach}, which studied online freelancers and projects, this paper focuses on long-term employment relationships. 
In contrast to the work of \citet{Oentaryo2018JobComposerLearning} and \citet{Guo2022IntelligentLearning}, which do not incorporate monetary rewards, the focus here is to chart the optimal path for the highest long-term income. 
Also, where \citet{Guo2022IntelligentLearning} posits any transition between jobs as possible, this study takes a more realistic approach and models transitions as a stochastic process learned from the data. 
\citet{Oentaryo2018JobComposerLearning} also model transitions as a stochastic process but assume it to be memoryless, making a person's next job dependent only on their current job. 
In contrast, this study introduces two settings: a \textit{naive setting} that makes the same assumption, and a \textit{standard setting} that leverages employees' past experiences to predict their next career move.

\section{The Proposed Career Path Recommendation Model}

We consider the problem of recommending a sequence of jobs --- a career path --- to the candidates that, if followed, would maximize their earnings during their foreseeable future.

Formally, given $C=\{c_1,...,c_n\}$ as $n$ candidates and $J=\{j_1,...,j_m\}$ as $m$ jobs, we define $R_c$ as the recommended career path generated for candidate $c$. We denote $W_c=\{w_{c,1},...,w_{c,k}\}$ as the work experience of candidate $c$, $k$ different jobs that $c$ worked in the past with $w_{c,1}$ being her first job and $w_{c,k}$ being her last (current) job. Each work experience contains information about the period (start date and end date) and the area (or role) that the candidate worked on in that job. The work area represents a high-level categorization for the jobs. In our experiments, we define it as a combination of an occupation and an industry. Examples are a Data Science role in Insurance or a Data Science role in Banking. Refer to Section~\ref{sec:datasets} for more details. 
We also denote $App_c=\{app_{c,j_1},...,app_{c,j_l}\}$ as $l$ jobs that candidate $c$ applied for in the past in which their outcomes are either hired or rejected. 
Finally, we denote $V$ as all the vacancies posted on the market. 

\begin{figure*}[!t]
\centering
\includegraphics[width=\textwidth]{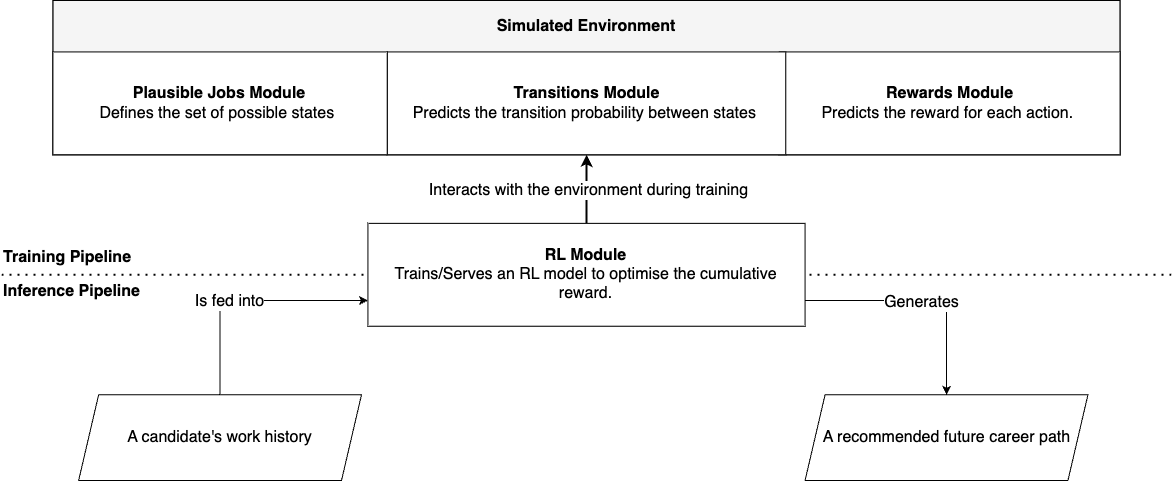}
\caption{The architecture of the proposed career path recommendation system.}
\label{fig:architecture}
\end{figure*}

Given these three input data (work experience $W$, job applications $App$, and vacancies $V$), our career path recommendation model comprises four distinct modules, as depicted in Figure~\ref{fig:architecture}. 

The first three modules—\textbf{Plausible Jobs}, \textbf{Transitions}, and \textbf{Rewards}—simulate the job market environment. 
The fourth module employs Reinforcement Learning (\textbf{RL}) to learn an optimal strategy for navigating these environments.

\paragraph{Plausible Jobs Module}
An employee's state at any given time is characterized by his current job, which is defined as a combination of occupation and industry, along with their work history. 
This concept forms the basis of the state space defined by this module, which comprises the set of available jobs and industries the agent can occupy. 
However, due to the constraints imposed by the dataset and to ensure a computationally feasible environment, our experiments are restricted to the 142 most prevalent jobs present in our dataset.

\paragraph{Transition Module}
Job applications do not always have deterministic outcomes; similarly, the actions taken by the agent should not always have deterministic outcomes. 
When the agent applies for a job and succeeds in being hired, it transitions from the current state $\mathbf{s}$ to a new state $\mathbf{s'}$. 
If unsuccessful, it remains in the current state $\mathbf{s}$. 
This transition occurs with probability $P(\mathbf{s'} | \mathbf{a}, \mathbf{s})$. 
A Random Forest binary classifier is trained on the Job Application data and is used to predict the aforementioned probabilities.
In other words, this module computes the transition probability between different jobs within the environment.

We consider the following approaches for computing the transition probabilities: 

\begin{itemize}
    \item \textbf{Last Job State Representation:} We assume that state $\mathbf{s'}$ only contains information about the last job of a person, implying that the probability of being hired depends solely on their latest job.
    \item \textbf{Full History State Representation:} Conversely, in the alternative approach, we assume that the state contains information about a person's entire work history. This second approach is closer to reality, but it also greatly increases the size and complexity of the state space, which could make learning more challenging and could potentially suffer from a lack of data.
\end{itemize}

\paragraph{Reward Module}
After each transition, this module is used to compute the reward earned from that transition. We define the reward in the form of the estimated salary that the individual earns after the transition. 
We use a Random Forest regressor trained on $V$ (i.e. all vacancies in the market) to predict the salary corresponding to each job. 
Given each $v \in V$ consists of the textual job description, and annual salary information, for our experiments we perform this prediction on a yearly basis for each pair of job role and industry. 
                       
\paragraph{Reinforcement Learning Module}
Lastly, the RL module uses RL algorithms to learn policies that can yield optimal rewards. After training, these models can be used to recommend high-income-yielding career paths to employees. We experiment with and compare multiple algorithms during the training of the RL module. The details of the algorithms are described in section \ref{sec_algs}.

\section{Methodology}

\subsection{Datasets}
\label{sec:datasets}
We conducted our experiments on anonymized data provided by Randstad as follows: 

\paragraph{Work Experience Dataset}
This tabular dataset consists of work experience items that employees may submit to Randstad either online or through consultants, or are directly taken from the administration of job placements made through Randstad. 
Relevant attributes for this research include: 1) Employee ID, 2) Job start and end dates, 3) ISCO code\footnote{\href{https://en.wikipedia.org/wiki/International_Standard_Classification_of_Occupations}{ISCO Wikipedia page: https://en.wikipedia.org/wiki/International\_Standard\_Classification\_of\_Occupations}} (occupation identifier), and 4) SBI code\footnote{ \href{https://www.kvk.nl/overzicht-standaard-bedrijfsindeling/}{SBI official website: https://www.kvk.nl/overzicht-standaard-bedrijfsindeling/}} (industry identifier).

Almost all the work experience items (99.99\%) pertain to Randstad placements, as these are jobs employees secured through Randstad. 
Most of the previous experiences (before using Randstad's services) are missing essential attributes. 

\paragraph{Vacancies Dataset}\label{sec:vacancies_dataset}
This dataset includes salary ranges for about six million vacancies, including their ISCO and SBI codes, posted on various Dutch websites. We use this dataset to estimate expected salaries for each occupation.

\paragraph{Job Applications Dataset}\label{sec:job_applications_dataset}
This dataset contains information on job applications made by candidates to Randstad's vacancies, with the outcome of each application (hired or rejected) also available.

\subsection{Data Preprocessing}
During preprocessing, we filtered out employees with missing data, jobs with durations less than a week, and employees with more than fifty work experience items from the work experience dataset, yielding 200K employees with 400K work experience items.

In line with Randstad's business model, most placements are short-term or temporary jobs common in staffing, resulting in a mean job duration of 161 days and a median duration of 95 days.

The average annual salary in the vacancies dataset is approximately 42K euros, with a median salary of 38K euros. 

\subsection{Reinforcement Learning Algorithms}\label{sec_algs}
Our experiments employ various Reinforcement Learning (RL) algorithms, which are primarily categorized into tabular methods and approximate RL methods.

\subsubsection{Tabular Methods}
Tabular methods are a class of RL algorithms that work well with a discrete, small state-action space. They maintain a table of values, with each entry in the table representing the value of each possible state-action pair.

\paragraph{State–action–reward–state–action (Sarsa)}
Introduced by \citet{rummery1994line}, Sarsa is an on-policy, tabular, temporal difference (TD) method. 
TD learning, which is a hybrid of Monte Carlo and dynamic-programming ideas, can learn directly from raw experience without a model of the environment's dynamics. 
Like dynamic programming, TD methods update estimates based in part on other learned estimates, without waiting for a final outcome (they bootstrap). 
The \textit{Sarsa} algorithm aims to learn an action-value function $q_\pi(s,a)$, providing the expected reward starting from state $s$, taking action $a$, and following the policy $\pi$.

\paragraph{Q-Learning}
Introduced by \citet{Watkins1992Q-learning}, Q-Learning is another tabular TD method. 
However, Q-Learning is an off-policy method, where the learned action-value function, Q, directly approximates $q_*$, the optimal action-value function, regardless of the policy followed (behavior policy).

\subsubsection{Approximate RL Methods}\label{sec:tabularvsapproximate}
While tabular methods perform well in environments with a small number of state-action pairs, they face challenges when the state-action space becomes large or continuous. They are not able to efficiently store the value of every possible state-action pair, nor can they generalize the value of unvisited state-action pairs effectively. This is where approximate RL methods come in to help with the Full History State Representation. These methods use function approximation, typically employing neural networks, to estimate the value of state-action pairs, allowing them to handle environments with larger or more complex state spaces more effectively.

\paragraph{Deep Q-Learning (DQN)}
DQN is an off-policy approach introduced by \citet{Mnih2013PlayingLearning}. DQN is the first successful deep-learning model to learn control policies directly from high-dimensional sensory input using reinforcement learning. It utilizes a convolutional neural network trained with a variant of Q-learning, taking raw pixels as input and estimating future rewards through a value function.

\paragraph{Advantage Actor-Critic (A2C)}
A2C is an approximate solution RL method that utilizes deep reinforcement learning for function approximation. Unlike DQN, A2C is an on-policy method. Introduced by \citet{Mnih2016AsynchronousLearning}, A2C is an actor-critic method, where the policy function is represented independently of the value function. The ``critic" model estimates the value function and the ``actor" learns the target policy. Both the Critic and Actor functions are parameterized with neural networks. As explained by \citet{Mnih2016AsynchronousLearning}, the main advantage of A2C over DQN is its faster training speed.

\subsection{Baselines}
Besides the above RL algorithms, we also perform experiments using two naive action selection approaches as baselines:

\paragraph{Greedy Most Common Transition} 
In this approach, the agent always applies for the job with the highest transition probability, that is the most likely job to be hired. In the case of multiple jobs with the same ranking, a random selection is made.
\paragraph{Greedy Highest Expected Reward} 
In this strategy, the agent applies for the job with the maximum expected reward, defined as the product of the transition probability and the immediate salary after the transition. In reality, this signifies the job with the highest likelihood of both being attained and yielding the highest immediate income. As before, in the case of multiple top-ranking jobs, a random selection is made.

\subsection{Evaluation Metrics}\label{sec:metrics}

We assess the effectiveness of our methods based on the income difference between \textit{observed career paths (factuals)} and \textit{recommended career paths (counterfactuals)}.
    
\paragraph{Observed Career Paths}
Using the Work Experience dataset, we generate a list of observed career paths and their corresponding income. 
Given that workers can hold multiple jobs simultaneously or have periods of unemployment, the dataset requires processing to align with the requirements of our simplified environment. Our models assume people only have one job at a time and there is no unemployment.
Therefore, in cases of simultaneous employment, we estimate each job's monthly salary and assume the worker earned the mean salary.
For periods of unemployment, we consider the salary from the worker's last job to be ongoing. 
    
\paragraph{Counterfactual Career Paths}\label{sec:counterfactuals}
After training each RL method, we sample observed career paths to generate their counterfactuals. 
These are the paths each model recommends, starting from the observed path's initial job, and lasting the same duration. 

\paragraph{Reported metrics}
For each model under consideration, we report two primary quantities - the \textit{Mean Factual} and \textit{Mean Counterfactual} accumulated rewards. 
These metrics represent the mean income accumulated by employees in reality versus the projected income they would have earned in a counterfactual scenario respectively. 

For an employee $e$ their factual income denoted as $FI$, over their career of $M$ months is calculated as 

\begin{equation}
    FI(e) = \sum^M_{m=1}{I(J_{e},m)}
\end{equation}

\noindent where $I(J_e,m)$ is a function that returns the salary that the employee $e$ had earned by performing job $J$ during the month $m$. Similarly, the counterfactual income is calculated as 

\begin{equation}
    CFI(e) = \sum^M_{m=1}{I(J^\prime_{e},m)}
\end{equation}

\noindent where $J^\prime$ is the job that employee $e$ would have performed if she had followed the recommendations of our system. 
Finally, the mean of these quantities is calculated over a sample of 20,000 observed (factual) and generated (counterfactual) career paths.

Following this, we present the \textit{Change \%}, illustrating the percentage change between the factual and counterfactual means. To determine the statistical significance of the observed difference, we calculate a \textit{p-value} using a two-sided permutation test with an alpha level of 0.05.

Furthermore, we detail the proportion of employees experiencing an income rise in the counterfactual world, referred to as \textit{Gainers}, along with the average magnitude of their income change. Similarly, we present data for those experiencing a decline, termed as \textit{Losers}, including the mean change in their income.

\subsection{Experimental results}\label{sec:baselines}
This subsection presents a detailed analysis of the experiment results. We juxtapose our factual and counterfactual career paths in terms of the mean income they generate. Additionally, we assess the effectiveness of our models by examining the percentage of gainers and losers as well as the magnitude of their respective income changes. 

\begin{table*}[h!]
\resizebox{\textwidth}{!}{%
\begin{tabular}{lrrrrrrrr}
\toprule
\rowcolor[HTML]{EFEFEF} 
Model & Mean $FI$ € & Mean $CFI$ € & Change \% & p-value & Gainers \% & Mean Gain \% & Losers \% & Mean Loss \% \\ \midrule
Baseline: Most Common & 90,283.42 & 89,644.81 & -0.7 & 0.69 & 8.85 & 8.17 & 11.62 & -8.40 \\ \midrule
Baseline: Highest Exp. Reward & 90,283.42 & 89,434.75 & -0.94 & 0.59 & 8.39 & 7.50 & 12.52 & -8.48 \\ \midrule
Q-Learning & 90,283.42 & 95,077.13 & \textbf{5.3} & 0.01 & 27.53 & 13.81 & 12.56 & -7.63 \\ \midrule
Sarsa & 90,283.42 & 94,836.08 & \textbf{5.04} & 0.01 & 32.84 & 11.50 & 10.95 & -7.46 \\ \bottomrule
\end{tabular}%
}
\caption{\textbf{Last Job State Representation}: Factual vs Counterfactual career paths. Metrics described in Section~\ref{sec:metrics}.}
\label{tab:naive_env:metrics}
\end{table*}

Table~\ref{tab:naive_env:metrics} presents the results for the Last Job State Representation, where the job seekers' state depends only on the last held job. 
From this table, we can observe that baselines do not perform significantly differently than the factual career paths, with differences under 1\% (at -0.7\% and -0.94\% for Most Common and Highest Expected Reward baselines, respectively). 
However, Q-Learning and Sarsa models perform well with a notable percentage of income gainers (27.53\% and 32.84\% respectively) and a reasonable mean gain percentage of 13.81\% and 11.5\% respectively.

\begin{table*}[h!]
\resizebox{\textwidth}{!}{%
\begin{tabular}{lrrrrrrrr}
\toprule
\rowcolor[HTML]{EFEFEF} 
Model & Mean $FI$ € & Mean $CFI$ € & Change \% & p-value & Gainers \% & Mean Gain \% & Losers \% & Mean Loss \% \\ \midrule
Baseline: Most Common & 90,386.16 & 95,871.78 & \textbf{6.18} & 0.02 & 71.07 & 17.27 & 25.15 & -13.39 \\ \midrule
Baseline: Highest Exp. Reward & 90,386.16 & 161,774.45 & \textbf{79.18} & 0.00 & 96.01 & 80.37 & 0.04 & -11.44 \\ \midrule
Deep Q-Learning & 90,283.42 & 94,547.94 & \textbf{4.7} & 0.01 & 67.91 & 16.95 & 27.87 & -13.71 \\ \midrule
A2C  & 90,283.42 & 95,616.29 & \textbf{5.9} & 0.00 & 70.82 & 17.22 & 25.35 & -13.64 \\ \bottomrule
\end{tabular}%
}
\caption{\textbf{Full History State Representation}: Factual vs Counterfactual career paths. Metrics described in Section~\ref{sec:metrics}.}
\label{tab:standard_env:metrics}
\end{table*}

Table~\ref{tab:standard_env:metrics} exhibits the outcomes for the Full History State Representation, where the state of a job seeker contains their full work history. 
The Highest Expected Reward baseline model stands out with a significant mean income change (79.18\%) and a large percentage of gainers (96.01\%). 
As we will discuss later, this is caused by Transitions module biases. Deep Q-Learning and A2C also show promising results but fail to outperform the baselines.


\begin{figure*}[h!]
        \centering
        \includegraphics[width=0.9\linewidth]{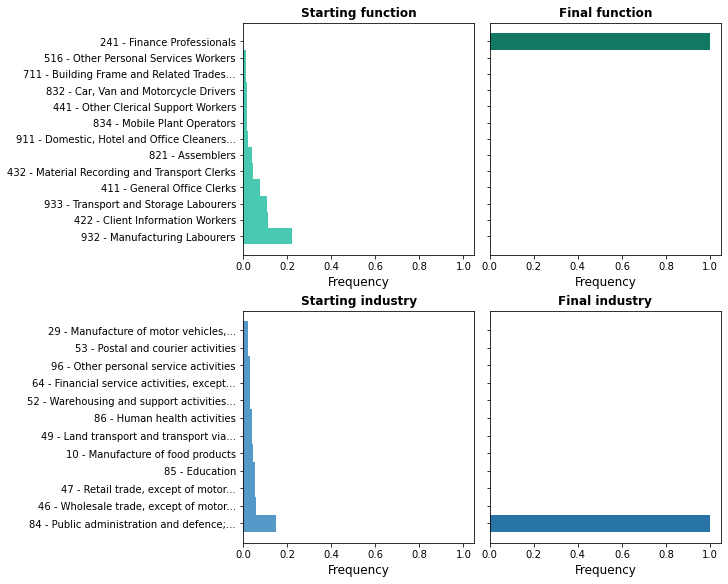}
        \caption{\textbf{Full History State Representation - Greedy Highest Expected Reward Baseline:} Starting and final distributions for the 12 most common functions and industries. The data were generated by running 1000 episodes of 40 time steps (10 years) each.}
        \label{fig:standard_env:highest_exp_reward:starting_final}
        \end{figure*}

\section{Results and Discussion}
In this chapter, we delve into the outcomes derived from our experimental setup featuring two unique versions of the transitions module - the Last Job State Representation and Full History State Representation. 
We start our discussion with findings from the two baseline methods outlined in Section~\ref{sec:baselines}. 
Subsequently, we elaborate on the results achieved through our efforts to learn an efficient policy. 
    
    \subsection{Implications of the Findings}
        
        \paragraph{Last Job State Representation}
        In the Last Job State Representation, our RL approaches learned policies that resulted in career paths with higher incomes than the observed career paths. In both the last job state representation and full history state representation, the learned policies improved the mean accumulated income by around $5\%$. 
        While not a drastic increase, this change is significant over longer time scales, such as an individual's career. 
        Notably, these improvements surpassed those of the baseline models. 
        However, there was also a significant amount, approximately 12\%, of agents for which the recommended paths performed worse than the observed.
        
        \paragraph{Full History State Representation}
        However, the Full History State Representation demonstrated a different pattern. While the DQN and A2C models also found policies improving counterfactual incomes, the baselines showed significantly larger improvements, particularly the \textit{Highest Expected Reward} baseline. This raises questions about the validity of the environment.
        After careful investigation, we found out that this environment can be easily exploited by the Highest Expected Reward baseline due to the small-but-substantial transition probabilities predicted by the Transitions module. As we can see in Figure~\ref{fig:standard_env:highest_exp_reward:starting_final}, regardless of the agent's starting state, it always applies and eventually succeeds to be hired for the highest-paying job of our dataset. By looking deeper into the classifier trained to predict the transition probabilities of the Full History State Representation, we see that regardless of the agents' prior experience there is always a small but significant probability of employment. 
        After analysis of the training data, we believe that this is due to missing data in the prior experiences of employees hired in senior positions. 
        Therefore, the training dataset is incomplete and depicts a world where someone can be hired, for example, as a Senior Finance professional with no experience in Finance.
        
        \paragraph{Comparison Between Environments}
        It's important to note that the results from the Last Job State Representation and Full History State Representation cannot be directly compared. Each model learns a policy to exploit the unique dynamics of the environment it is trained on, therefore, the ground truth differs for each environment. As such, results should be compared within the specific environment they originated from.
    
    \subsection{Limitations of the Study}
    
        \paragraph{Job Filtering}
        The experiments relied on a narrowed field of 142 most common jobs. The decision to do so was driven by the challenges posed by the vast state space and unreliable transition probability prediction for less common jobs. 
        
        \paragraph{The Cost of an Action}
        The research assumes no monetary cost for applying to a job, which is typically not the case in reality, where applications cost both time in interviewing or preparing, and perhaps other forms of preparation (e.g., studying). Considering the real-world costs in future studies could bring the environment formulation closer to reality.
        
        \paragraph{Continuous Employment}
        The assumption that individuals are always working doesn't account for potential breaks in employment. These breaks could result from various factors, including vacations, relocation, or further education, and should be considered in a more realistic formulation of the job market as an MDP.

        \paragraph{Rivalrous market}
        Another notable limitation is that the real-world job market is inherently rivalrous --- a job offered to one applicant becomes unavailable to others. Recommending the most highly paying jobs to all users could potentially lead to an overwhelming influx of applications for those positions, resulting in many disappointed users due to the increased competition. Our approach focuses on income optimization, but it's crucial to recognize that a well-balanced approach considering job availability, individual preferences, and market dynamics is necessary to avoid an undue concentration of applications in specific roles. Future research should delve into strategies that account for these challenges while still aiming for income optimization to create a more comprehensive and realistic career planning model.
        
        \paragraph{State Space and the Markov Property}
        The models used in this research made different assumptions about state space and respected the Markov Property in different ways. The Last Job State Representation model simplified the state to be a job, assuming an employee's future depends solely on their last job. On the other hand, the Full History State Representation model considered employees' whole work experience as part of the state. The latter approach is closer to reality but can create states with too many dimensions, slowing policy learning. An option suggested in the literature for similar challenges is learning low-dimensional embeddings and reducing the state space size.
    
\section{Conclusion}
        In conclusion, this research explored the use of artificial intelligence, specifically reinforcement learning, in the field of career planning. By harnessing data on employee work experience and job applications, the research aimed to recommend career paths that maximize long-term income for individuals.
        
        The findings of this study showed promising results in both the Last Job State Representation and Full History State Representation approaches. The reinforcement learning models, particularly Q-Learning and Sarsa, were able to learn policies that improved the counterfactual incomes of individuals. In the Last Job State Representation, the mean accumulated income increased by around 5\%, surpassing the performance of the baseline models. 
        
        However, it is important to acknowledge the limitations and failures of the Full History State Representation. The baseline models exhibited greater improvements in counterfactual incomes compared to the reinforcement learning models. This discrepancy was due to inaccuracies in the transition probability predictions, which allowed the Highest Expected Reward baseline to exploit the system.
        
        These limitations indicate the need for further research to refine and improve the Full History State Representation. Future studies could explore alternative methods for estimating transition probabilities and address the issue of missing prior experience data. By addressing these challenges, future research can work towards creating more robust and accurate environments that better reflect the complexities of real-world career planning.


\balance
\bibliography{references}

\end{document}